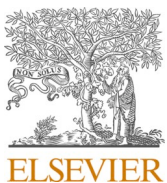
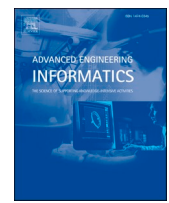

Full length article

# Large language model for patent concept generation

Runtao Ren [a,*], Jian Ma [a], Jianxi Luo [b,*]

[a] *Department of Information Systems, City University of Hong Kong, Kowloon Tong, Hong Kong*
[b] *Department of Systems Engineering, City University of Hong Kong, Kowloon Tong, Hong Kong*

A B S T R A C T

In traditional innovation practices, concept and IP generation are often iteratively integrated. Both processes demand an intricate understanding of advanced technical domain knowledge. Existing large language models (LLMs), while possessing massive pre-trained knowledge, often fall short in the innovative concept generation due to a lack of specialized knowledge necessary for the generation. To bridge this critical gap, we propose a novel knowledge finetuning (KFT) framework to endow LLM-based AI with the ability to autonomously mine, understand, and apply domain-specific knowledge and concepts for invention generation, i.e., concept and patent generation together. Our proposed PatentGPT integrates knowledge injection pre-training (KPT), domain-specific supervised finetuning (SFT), and reinforcement learning from human feedback (RLHF). Extensive evaluation shows that PatentGPT significantly outperforms the state-of-the-art models on patent-related benchmark tests. Our method not only provides new insights into data-driven innovation but also paves a new path to fine-tune LLMs for applications in the context of technology. We also discuss the managerial and policy implications of AI-generating inventions in the future.

## 1. Introduction

In today's fast-paced scientific era, the ability to convert ideas and knowledge into intellectual property (IP) is crucial for every organization and nation. IP not only provides legal protection but also plays a vital role in transforming groundbreaking scientific discoveries into tangible innovations that drive economic and technological progress [1]. The World Intellectual Property Organization (WIPO) Report analyzed nearly 40 million patent applications, over 70 million scientific publications, and more than $300 trillion in goods and services involved in export-driven economic activities [2]. Over the past two decades, the top eight countries in patent ownership have accounted for 50 % of global exports, 60 % of scientific publications, and 80 % of international patent grants [2]. This indicates that industries with a stronger IP advantage are more likely to win in the innovation race. For example, although Nokia sold its mobile phone business to Microsoft in 2013, any manufacturer that uses Nokia's patents in their 5G phones must still pay Nokia a €3 royalty fee per device [3]. Over the ensuing decade, Nokia not only maintained its market presence but also continued to actively participate in the telecommunications sector. This demonstrates that for companies within the industrial sector, IP is a vital asset for strengthening market position. Thus, the ability to generate novel and useful concepts and transform them into IP is critically important.

Data-Driven Innovation (DDI) can uncover new insights and potential directions for invention by analyzing and mining concepts from vast amounts of data [4]. Particularly, integrating Language Models (LM) and Natural Language Processing (NLP) into patent analysis can significantly advance innovation efforts. Within the contexts of manufacturing and Internet of Things (IoT) technologies, research has demonstrated how LMs can enhance invention by automatically extracting meaningful concepts from extensive technical documents and patents [5,6]. For instance, the LM with fine-tuning (e.g., BERT) can mine and identify key innovative concepts from extensive scientific texts and visualize these technological innovations in a manner that is easily comprehensible to decision-makers [7]. This process, known as concept space discovery, enables companies eager to leverage emerging technologies and market shifts to swiftly capture the cutting edge of technological innovation.

Despite the advancements in LMs greatly facilitating the identification of conceptual entities and relationships, their application in the precise and highly regulated fields of concept and patent generation presents a series of challenges. Traditional patent drafting requires preliminary technical concepts and specialized expertise. Given the complexity of patent documents and the stringent legal requirements, even experienced patent attorneys may struggle to ensure the speed and quality of patent applications [8]. Although LMs demonstrate remarkable capabilities in processing textual data, encoding-based LMs (e.g., BERT) can only map concepts to a vector space and cannot present specific concepts in an intuitive text form to individuals [9]. At the same time, due to limitations in parameters and input token length, traditional LMs may not effectively capture the extensive text found in patents [10]. Consequently, it may be worthwhile to consider deploying large language models (LLMs) based on decoding to address these issues. However, the writing of patent documents requires accurate and clear clarification of the unique innovation of the invention concept to meet strict legal standards [11]. Hence, applying LLMs in the generation of initial concept text that can be converted into a patent still presents significant difficulties [12].

The current dilemma is not merely the automatic generation of innovative concepts, but rather leveraging the power of AI to unlock the potential for creating vast amounts of artifacts. Therefore, we require a system that can comprehend complex scientific knowledge, extract valuable concepts, and seamlessly transform these concepts and knowledge into inventions, as shown in Fig. 1. A visionary goal of DDI is that, someday, artifacts will mine and identify concepts and then articulate these as legally valid and technically precise inventions [13]. Consequently, the following pivotal research question is proposed:

**How can AI be harnessed to generate innovative concepts for facilitating invention?**

This research aims to integrate AI into the core of concept discovery and knowledge utilization. The ultimate objective is to develop an artifact capable of reading and comprehending scientific literature, extracting underlying concepts and knowledge, and autonomously designing initial concepts of inventions that are both legally sound and technically precise. To achieve this objective, we have developed a Knowledge Fine-Tuning (KFT) framework for technical texts aimed at enhancing the performance of LLMs in generating new initial concepts of patents, exemplified through our instantiation of PatentGPT.

Specifically, PatentGPT begins by automatically mining entities and relationships from patent texts to build the Knowledge Graph (KG), which is then utilized as a hybrid input for further pre-training (PT) of the LLM. This is followed by Supervised Fine-Tuning (SFT) based on invention generation scenarios, enabling the LLM to acquire IP knowledge in specific contexts. Finally, the model is refined through Reinforcement Learning from Human Feedback (RLHF) to ensure that the generated content aligns with human knowledge recognition, thereby enhancing the LLM's ability. To distinguish our work from previous studies, our research provides the following contributions:

(1) We introduced a novel KFT workflow that enhances LLM performance in concepts and patents generation by incorporating the method of knowledge injection, which demonstrated that the continuous Knowledge Injection Pre-training (KPT) process significantly improves the model's learning capabilities during subsequent fine-tuning phases compared to traditional continuous PT.

(2) The PatentGPT is the first system specifically tailored for the initial concept generation of patents suitable for U.S. patent application, designed to assist researchers and junior engineers across various industries in inventing.

(3) Compared to existing LLMs, our model achieved state-of-the-art performance in related benchmark testing and demonstrated that training with the KFT method significantly enhances model capabilities in specialized domains, with only a minimal sacrifice in general abilities.

## 2. Literature review and background

### 2.1. Data-driven innovation for concept generation

Concept-Knowledge Theory (C-K Theory) provides a theoretical foundation to explain how inventions originate [14]. According to this theory, concepts are potential or existing terms, while knowledge consists of validated axioms that can support these concepts. Under C-K Theory, the process of invention can be viewed as the creation of new concepts through the combination of various knowledge or conceptions, as illustrated in Fig. 2 (e.g., blockchain is a new concept synthesized from existing cryptographic and distributed storage technologies). Conversely, patents can also be seen as new knowledge generated through the formalization of innovative concepts (e.g., the invention of the internal combustion engine emerged through the innovation of dynamics knowledge). Thus, in the design process of artifacts, knowledge and concepts are interwoven and transformed, mutually fostering innovation [15].

Based on this interaction, the myriad concepts and knowledge entities in inventions can be viewed as potential resources for DDI. In the past, research has predominantly employed encoding-based methods to represent these knowledge and concepts, depicting them as entities within a vector space. For instance, Jiang et al. utilized the LMs to examine the dependency networks of patent technology elements, thereby aiding in the discovery of potential innovative pathways [16]. Similarly, Geum and Kim used dynamic network analysis combined with the LDA model to identify the trend of technological integration [17]. While these methods provide valuable insights when mapping concepts to vector spaces, they are primarily used for interpreting and analyzing existing data, rather than directly facilitating the generation of new concepts and invention.

Given these limitations, there is an urgent need for an artifact that can effectively leverage knowledge and concepts to directly support and enhance the invention process. Through extensive PT and SFT, LLMs are able to learn the complex linguistic features and technical details of patent documents, surpassing the limitations of traditional methods [18]. Moreover, the generative capabilities of LLMs can also aid in constructing innovation points of patent from scratch, thereby better supporting creative thinking and the invention process, rather than merely analyzing and predicting. Hence, LLMs offer significant advantages over traditional LMs.

### 2.2. Fine-tuning method for patenting

Although there are efforts to employ LLMs for generation tasks within specific domains, these methods often fall short in the highly structured area of patent writing. To effectively address the issues of automated text generation, especially in niche domains, LLMs need to possess prior knowledge relevant to those fields [19]. Previous literature has explored the fine-tuning method to enhance the capabilities of LLM in new technical concepts generation, as shown in Table 1. For these fine-tuning-based models, the common approach is to select a base model and then construct a dialogue dataset for SFT. SFT is an approach for fine-tuning models on domain-specific labeled datasets. While SFT can help models learn the specific style and requirements of patent writing, it typically requires a large amount of annotated data and still struggles to generalize beyond the examples it was trained on.

In the earliest study about generating original design concepts, Trappey et al. used sequence to sequence with attention models to develop an intelligent patent summarization system [20]. Their model combines extraction and generative abstract methods, and has been studied and validated through case studies in the field of intelligent machinery patents. However, traditional NLP techniques are unable to handle long texts and complex contexts like large models. Later, Zhu and Luo demonstrated the ability of GPT-2 and GPT-3 to integrate near-field and far-field knowledge sources for early-stage design tasks through a combination of fine-tuning and few-shot learning methods [21]. Another study by Zhu and Luo used patent text data to fine-tune GPT-2 for generating design concepts [22]. Subsequently, to explore the potential of LLMs in assisting humans with concept design across different fields, they also used fine-tuning as an approach for the few-shot learning of GPT-3 [23].

In the literature on writing concept texts for patents, Lee discussed the performance of LLMs in generating patent abstracts and analyzed the quality and originality of the generated content [24]. However, this article did not provide a method for training LLMs, but instead tested the performance of different models on IP-related tasks. MoZI represents the first study to explore how LLMs can empower IP rights. Ni et al. published multiple benchmarks for IP-related tasks and demonstrated the benefits of fine-tuning LLMs using IP-specific data [25]. Although the model developed by Ni et al. was fine-tuned to generate patentable concept text, it seems to have lost a significant amount of general knowledge, and over-reliance on a large volume of specialized data limits its scalability. This highlights the challenge of injecting specific knowledge into the model while balancing it with the broader knowledge base, an issue that requires further attention.

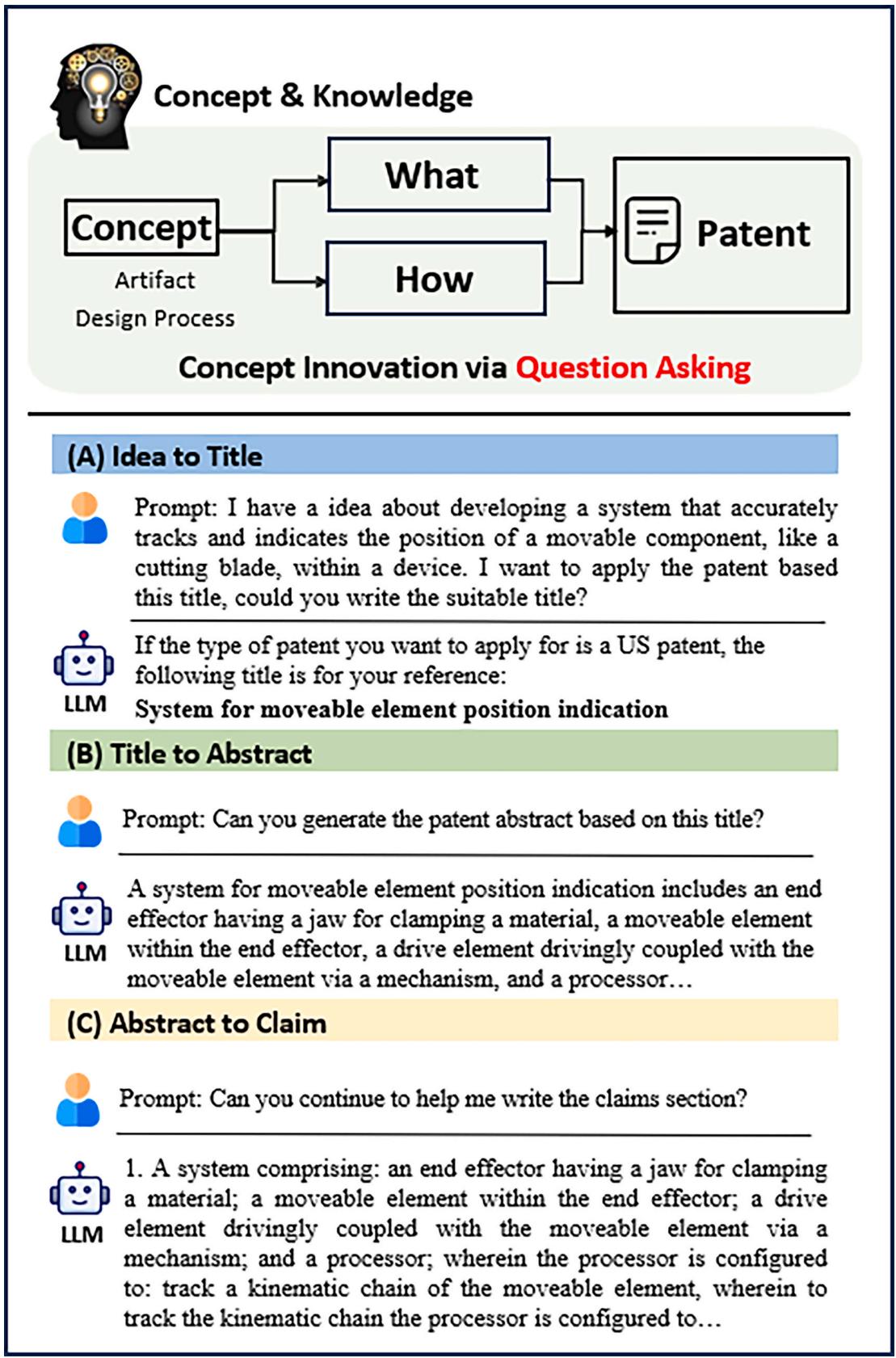

**Fig. 1.** Diagram of concepts turning into patents.

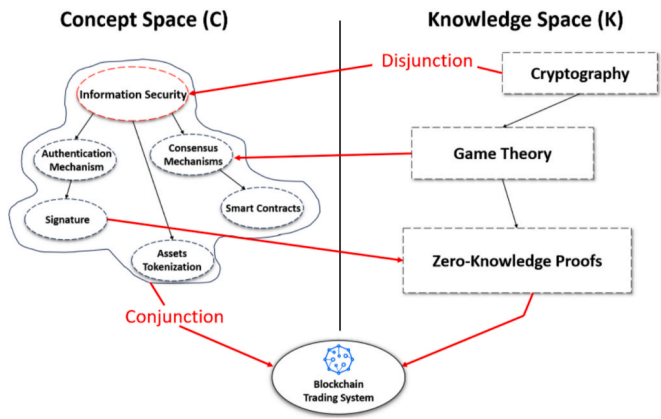

**Fig. 2.** Concept generation created by C-K Theory.

**Table 1**
Representative previous studies of LLM and LM in IP-related field.

| Year | Author | Method | Advantage | Limitation |
|---|---|---|---|---|
| 2024 | Jiang et al., [16] | Embedding | Developing a system for patent knowledge graph using LMs to aid in the application of innovative design knowledge | This structured approach may not be easily adaptable to rapidly changing fields |
| 2020 | Geum and Kim [17] | Network Analysis | Dynamic network analysis combined with the LDA model is introduced to the patent field for understanding technology trends | The LDA model might not capture the full semantic richness of technical documents |
| 2020 | Trappey et al., [20] | Finetuning | Compared to LLMs, it has lower operating costs and resource consumption | Unable to handle long texts and complex contexts like LLMs |
| 2022 | Zhu and Luo [21] | Finetuning | Show the potential of LLMs to augment human creativity in design by leveraging external knowledge sources systematically | It does not address the structured and highly technical aspects required in patent writing |
| 2022 | Zhu and Luo [22] | Finetuning | Explore the LLMs in design concept generation, showing that LLMs can handle problem-driven and analogy-driven reasoning tasks effectively | It does not dive into the domain-specific knowledge necessary for IP documentation |
| 2023 | Zhu and Luo [23] | Finetuning | Show how AI can augment the early design process and collaborate with designers in brainstorming sessions. | Its generative ability and adaptability may be limited when applied to broader technological innovations |
| 2024 | Jieh-Sheng Lee [24] | SFT | Focus on domain-specific training for improved accuracy in patent texts | Lacks adaptability to evolving patent drafting standards |
| 2024 | Ni et al., [25] | PT, SFT | Design a multilingual benchmark for large language models in IP | Single training method cannot guarantee high-quality patent text writing |

From the literature review above, we can identify existing research gaps: current methods are unable to fully integrate domain-specific knowledge with textual data to enhance the learning process. Existing approaches treat concepts or knowledge and text as separate entities, failing to create a holistic learning environment that allows the model to utilize both simultaneously. This gap necessitates a new approach that combines structured knowledge with original text to strengthen the training process.

In response to these gaps, the KFT framework was proposed. This framework is designed to bridge the gap between textual and knowledge-based learning by:

(1) **Knowledge Injection**: Transforming structured triples from patent databases into text and injecting this knowledge into the model during the pre-training phase, enabling the model to learn the requisite knowledge and context for patent concept writing.

(2) **SFT with Domain-Specific Datasets**: Enhancing the model's ability to perform specific patent writing tasks by fine-tuning it with carefully curated patent-specific datasets.

(3) **RLHF**: Further refining the model to ensure the generated content is both innovative and adheres to human standards of knowledge recognition, thereby continuously improving the model through a feedback loop.

## 3. Method

Our KFT framework systematically integrates LLMs with KGs for fine-tuning. The approach comprises four main stages: *knowledge extracting, knowledge injecting, knowledge learning,* and *knowledge feedbacking*. As shown in Fig. 3, LLM first extracts knowledge from patent text to construct knowledge graphs, then converts them into coherent training text and uses the converted text and the initial patent text as training data for knowledge injecting. Then, the model is trained on conversational data using SFT. Finally, RLHF refines outputs using feedback to enhance accuracy and relevance.

### 3.1. Knowledge extracting

In the initial stage, we focus on the general corpus $G = \{g_1, g_2, ..., g_n\}$ from patent textual data of USPTO and employ the LLM $\pi_\theta$ to automatically extract concept entities and relationships using the prompt for generating KGs (shown in Appendix A) based on the method of Luan et al. [26] for constructing the knowledge graph $K$ with triples $(h, r, t)$ as below.

**Step 1** Knowledge Extracting

1: **Input**: General corpus $G = \{g_1, g_2, ..., g_n\}$
2: **Output**: Knowledge graph $K = \{h, r, t\}$
3: Knowledge graph $K = [\,]$
4: **For** $g_i$ in $G$:
5:     $T_i = \mathrm{argmax}_{(h, r, t)}\, \pi_\theta\,(h, r, t \mid g_i)$
6:     // $h$ is the head entity
7:     // $r$ is the relationship
8:     // $t$ is the tail entity
9:     // $\theta$ is the prompt for extracting triplet
10: **Aggregate** each $T_i$ to construct knowledge graph $K$:
11:     $K = \bigcup_{i=1}^{n} T_i$
12: **Return** Knowledge graph $K$;

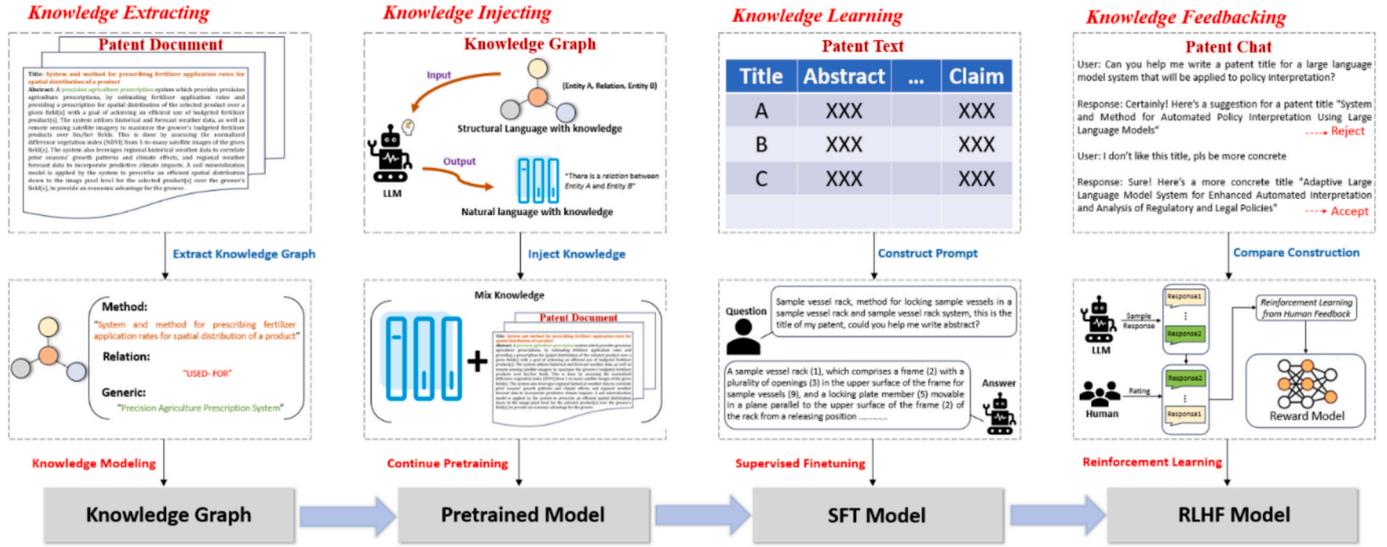

**Fig. 3.** Framework of knowledge finetuning.

### 3.2. Knowledge injecting

Following the extraction of knowledge, each structured triple in extracted knowledge graph $K$ is transformed into natural language sentences $s_i$ using the prompt shown in Appendix B and stored each $s_i$ in knowledge corpus $S$. This conversion process can transform structured triplet text into unstructured natural text, enabling LLMs to better learn the knowledge and concepts from patent texts during pre-training. Finally, the LLM undergoes continued pre-training using both knowledge corpus $S$ and general corpus $G$ for fine-tuning.

---

**Step 2** Knowledge Injecting

1: **Input**: $S$ and $G$
2: **Output**: Fine-tuned model
3: *Knowledge corpus $S$* = [ ]
4: **For** $T_i$ **in** $K$:
5:     $s_i = \pi_\omega (h_i, r_i, t_i)$
6:     // $\omega$ is the prompt for converting sentences
7: **Aggregate** each $s_i$ to get knowledge corpus $S$:
8: $L_{PT}(\chi) = \alpha E_{s \in S}[\log \pi_\chi(y \mid s)] + \beta E_{g \in G}[\log \pi_\chi(y \mid g)]$
9: // $L_{PT}$ is the loss function for pre-training
10: // $\chi$ is the parameters of LLM
11: // $\alpha$ and $\beta$ are the parameters of importance
12: // $L_{PT}$ is the loss function for pre-training
13: **Return** Fine-tuned model $\pi_\chi$;

---

### 3.3. Knowledge learning

After obtaining the model with knowledge injection by continued PT, the SFT is used to align the outputs of the LLM with specific tasks by training it on dialogue datasets $Q = \{ (q_i, a_i), (q_2, a_2), …, (q_n, a_n) \}$. To convert a general LLM with domain knowledge, it is not enough to simply inject background concepts or knowledge into the model. It is also necessary to learn how to let the model use knowledge and concepts according to specific scenarios. Therefore, this step is called the *knowledge learning* process. This procedure reduces the difference between the anticipated outputs and the ground truth labels using SFT method, which aids the model in producing more accurate and contextually relevant replies in the concrete field.

---

**Step 3** Knowledge Learning

1: **Input**: $Q = \{ (q_1, a_1), (q_2, a_2), …, (q_n, a_n) \}$
2: **Output**: Fine-tuned model
3: $L_{SFT}(\chi) = E_{(q,a) \in Q}[\log \pi_\chi(a \mid q)]$
4: **Return** Fine-tuned model $\pi_\chi$;

---

### 3.4. Knowledge feedbacking

In this phase, the model is improved through RLHF. Specifically, we collect user feedback when deploying the model, test the user's preference for the same question, and mark the answer as human preference (i.e., consistent with human knowledge recognition), so this process is called *knowledge feedbacking*. Then these labeled human feedback datasets, denoted as $H = \{ (q_1, p_1, n_1), …, (q_n, p_n, n_n) \}$, are used train a reward model. Here, $c$ represents a query, $p_i$ denotes the preferred response, and $n_i$ denotes the non-preferred response. To stabilize reinforcement learning, we employ proximal policy optimization (PPO) with a reward signal provided by the reward model score. The reward model is trained to predict human preferences by minimizing the loss function $L_{RM} = -E_{(q,p,n) \in H}[\log \sigma(r_\rho(p \mid q) - r_\rho(n \mid q))]$, where $r_\rho(p \mid q)$ and $r_\rho(n \mid q)$ are the reward scores assigned by the reward model to the preferred and non-preferred responses, respectively, and $\sigma$ is the sigmoid function. Finally, the LLM is fine-tuned using PPO, with the reward model providing feedback on the quality of generated responses. The training process involves iteratively updating the model's policy to maximize the expected reward while maintaining stability through PPO's clipping mechanism.

---

**Step 4** Knowledge Feedbacking

1: **Input**: $H = \{ (q_1, p_1, n_1), …, (q_n, p_n, n_n) \}$
2: **Output**: Fine-tuned model
3: $L_{RM} = -E_{(q,p,n) \in H}[\log \sigma(r_\rho(p \mid q) - r_\rho(n \mid q))]$
4: $L_{PPO} = E_{(q,p,n) \in H}[\min(r_t(\chi)A_t, \text{clip}(r_t(\chi), 1-\varepsilon, 1+\varepsilon)A_t)]$
5: $r_t(\chi) = \pi_\chi(a \mid q) / \pi_{\chi old}(a \mid q)$
6: // $r_t(\chi)$ is the ratio between current and old strategy
7: // $A_t$ is the advantage estimate
8: **Return** Fine-tuned model $\pi_\chi$;

---

## 4. Experiments

### 4.1. Experiment setup

We chose the Qwen2-1.5B as the base model for conducting KFT [30]. The Qwen2-1.5B model is a 1.5 billion-parameter language model developed by Alibaba's DAMO Academy as part of the Qwen2 series. Qwen2-1.5B model achieves a balance between computational efficiency and performance, allowing for effective fine-tuning even on limited hardware resources. The selected Qwen2-1.5B is an instruction-tuned variant specifically optimized for task-specific performance, particularly in natural language understanding and generation tasks that require precise adherence to complex instructions.

The fine-tuning and inference processes are conducted on NVIDIA 4090 GPUs. During fine-tuning, we adopt the LoRA method, batch size of 1, learning rate of 5e-5, and training epochs of 3. For inference, we set the temperature to 0.7 on the LLaMA-Factory and lm-evaluation-harness framework [27,28].

### 4.2. Data

#### 4.2.1. USPTO patent text dataset

This dataset was collected on the USPTO website, covering patents published between 2021 and 2022 according to the International Patent Classification (IPC) [29]. Specifically, the IPC organizes patents into eight main sections, which encompass a diverse range of industries and technologies: A: Human Necessities, B: Performing Operations; Transporting, C: Chemistry; Metallurgy, D: Textiles; Paper, E: Fixed Constructions, F: Mechanical Engineering; Lighting; Heating; Weapons; Blasting, G: Physics, H: Electricity. To ensure a balanced representation of industries, we collected approximately 20,000 patents, with each IPC section containing between 2,000 and 4,000 patents. The dataset includes patent titles, abstracts, and claims, which are used for pre-training.

#### 4.2.2. Patent knowledge graph dataset

To enhance the model's conceptual understanding and knowledge comprehension capabilities, we constructed a patent knowledge graph (KG) dataset based on the method proposed by Luan et al. [26]. The KG dataset is built using a subset of the USPTO Patent Text Dataset, specifically 30 % of the total patents. Specifically, we use LLM to automatically extract entities and relationships by utilizing a carefully designed prompt (as shown in Appendix A). The LLM identifies and classifies entities and relationships strictly according to the predefined types outlined in Table 2 (entity types) and Table 3 (relationship types). The final KG dataset contains approximately 100,000 entities and 300,000 relationships. The extracted entities and relationships are represented as triples (e.g., <entity1, relationship, entity2 > ), which are then converted into natural language descriptions using the prompt as shown in Appendix B for knowledge injecting.

#### 4.2.3. SFT dialogue dataset

We have created a dialogue dataset consisting of patent drafting dialogues to further fine-tune (i.e., SFT) the interactive functionality of the model. This dataset is constructed based on the USPTO's patent text dataset, with a total of 60,000 question–answer pairs related to drafting patent titles, abstracts, and claims.

#### 4.2.4. Human feedback dataset

To improve the quality and legal accuracy of the model's responses, we curated a human feedback Q&A dataset containing 2,000 interactions. This dataset is based on the scenarios covered in the SFT Dialogue Dataset, focusing on generating patent titles, abstracts, and claims. For each query, we provide two candidate answers: one preferred answer and one non-preferred answer. The preferred answers are annotated by experienced patent practitioners to ensure high-quality feedback for model training.

**Table 2**
Entity category in KG.

| Entity Type | Definition |
|---|---|
| Task | Applications, problems to solve, systems to construct |
| Method | Methods, models, systems to use, or tools, components of a system, frameworks |
| Material | Data, datasets, resources, Corpus, Knowledge base |
| Generic | General terms or pronouns that may refer to entity but are not themselves informative, often used as connection words |

**Table 3**
Relation category in KG.

| Relation Type | Definition |
|---|---|
| Used-for | B is used for A, B models A, A is trained on B, B exploits A, A is based on B |
| Feature-of | B belongs to A, B is a feature of A, B is under A domain |
| Hyponym-of | B is a hyponym of A, B is a type of A |
| Part-of | B is a part of A |
| Compare | Symmetric relation. Opposite of conjunction, compare two models/methods, or listing two opposing entities |
| Conjunction | Symmetric relation. Function as similar role or use/incorporate with |

### 4.3. Baselines

We selected a variety of models with different scales and architectures for comparison, including Qwen2-1.5B [30], Llama-2 [31], Llama-3.2 [32], phi-2 [33], and ChatGPT-4o [34], Gemini 1.5 Flash-8b [35], Claude2 [36]. We also compared the PatentGPT against the IP domain's state-of-the-art model MoZI-7b [25] to demonstrate the performance of our model in the invention field.

### 4.4. Metrics

**Automated Evaluation**. In order to comprehensively evaluate the performance of the concepts and patents generation, we used a variety of language generation quality evaluation indicators. These indicators include ROUGE-1, ROUGE-2, ROUGE-L, BLEU-4, and BERTScore, which together measure the quality of the text generated by the model from different dimensions [37–39]. The following are the definitions and formulas:

(1) **ROUGE-1**: It measures the degree of overlap between the generated text and the 1-gram in the reference text, emphasizing word-level coverage. Its formula is:

$$\text{ROUGE-1} = \frac{\sum_{w \in \text{Ref}} \text{Count}_{1\text{-gram}}(w)}{\sum_{w \in \text{Ref}} \text{Count}(w)}$$

where $\text{Count}_{1\text{-gram}}(w)$ is the number of 1-gram matches between the generated text and the reference text and $\text{Count}(w)$ is the total number of 1-grams in the reference text

(2) **ROUGE-2**: This indicator measures the degree of overlap between the generated text and the 2-gram in the reference text, and is suitable for capturing more complex phrase matches. The following is the formula corresponding to this indicator:

$$\text{ROUGE-2} = \frac{\sum_{w \in \text{Ref}} \text{Count}_{2\text{-gram}}(w)}{\sum_{w \in \text{Ref}} \text{Count}(w)}$$

where $\text{Count}_{2\text{-gram}}(w)$ is the number of 2-gram matches between the generated text and the reference text and $\text{Count}(w)$ is the total number of 2-grams in the reference text.

(3) **ROUGE-L**: This indicator is based on the longest common subsequence (LCS) to calculate F-measure, which reflects the similarity in semantic order and structure between the generated text and the reference text. Its formula is:

$$R_{LCS} = \frac{LCS(X,Y)}{m}$$

$$P_{LCS} = \frac{LCS(X,Y)}{n}$$

$$F_{LCS} = \frac{(1+\beta^2)R_{LCS}P_{LCS}}{\beta^2 R_{LCS} + P_{LCS}}$$

where $LCS(X, Y)$ is the longest common subsequence length for generating text $X$ and reference text $Y$, $m$ is the length of reference text, $n$ is the length of generated text, and $\beta$ is the ratio of precision to recall.

(4) **BLEU-4**: BLEU mainly focuses on the n-gram accuracy of generated text, and also reflects semantics and fluency. For the generating task of LLMs, BLEU-4 can be used to measure the accuracy of phrase matching in the generated text. The following is the formulation of this metric:

$$\text{BLEU} = \text{BP} \cdot \exp\left(\sum_{n=1}^{N} w_n \cdot \log p_n\right), \text{BP} = \begin{cases} 1 & \text{if } c > r \\ e^{(1-r/c)} & \text{if } c \leq r \end{cases}$$

where $w_n$ is the weight of n-grams, $m$ is the length of reference text, $BP$ is the length penalty factor, $c$ is the length of the generated text, and $r$ is the length of the reference text.

(5) **BERTScore**: It is a semantic similarity-based evaluation metric for generated text, which uses pre-trained LMs (e.g., BERT) to generate word embeddings and calculates the semantic match between the generated text and the reference text through cosine similarity. The following is the formulation of this metric:

$$\text{BERTScore} = \frac{1}{|T|} \sum_{t_i \in T} \max_{r_j \in R} \text{cosine}(E(t_i), E(r_j))$$

where $t_i$ is the word in the generated text, $r_j$ is the word in the reference text, $E(t_i)$ represents word embedding using LM, $\text{cosine}(E(t_i), E(r_j))$ is the cosine similarity between generated words and reference words, $T$ and $R$ are the sets of words in the generated text and reference text, respectively.

**IP_Quiz**. It is sourced from a public dataset designed to evaluate a model's understanding of IP concepts and regulations [25]. For each question, the model is required to select the correct answer from a list of candidates. The number of correct selections indicates the model's proficiency in comprehending IP.

**IP_Match**. This benchmark also comes from the public benchmark test set on Hugging Face and is used to evaluate whether the model truly understands the invention described in the patent document and whether it can accurately distinguish different patents [25].

**IP_Exam**. This is a custom benchmark derived from multiple-choice Q&A content used in the U.S. Patent Bar Examination [40]. A higher performance on this benchmark indicates a model's capability to not only understand but also accurately apply IP law principles.

**MMLU**. Massive Multitask Language Understanding (MMLU) is a widely recognized benchmark designed to assess a model's performance across a diverse set of tasks, encompassing a broad range of subjects from the humanities, sciences, and professional disciplines [41]. A higher score on the MMLU benchmark reflects the model's versatility and robustness in handling a wide array of language understanding tasks.

**Concept Generation Evaluation**. To measure the content created by artificial products, we adopt a two-stage approach to measure *reasonability* and *innovation*. To assess the reasonability of the generated initial concept texts of patents, we adopted a general approach by using advanced LLMs as evaluators [42]. Specifically, we randomly selected 100 samples from the test set and employed the ChatGPT4 to act as a judge. The task of LLM is to determine whether each generated initial concept text meets the reasonability of the patent application. For measuring the innovation of the generated patent text, we filtered out the texts that were deemed reliable (i.e., conforming to USPTO standards) and then assessed their innovation based on the evaluation method by TechNet [43]. The innovation score is calculated by analyzing the novelty and uniqueness of the content in relation to existing patent databases. A higher innovation score indicates that the generated concept texts contain more innovation.

## 5. Benchmark testing

### 5.1. Patent writing

The patent writing benchmark is designed to evaluate the ability of LLMs to write patentable concept texts. The test set includes 1,000 instances sampled from all eight major IPC (International Patent Classification) categories, covering a wide range of industries such as chemistry, mechanical engineering, electronics, and biotechnology. The designed prompt used for generating the patent texts is provided in the Appendix C.

The results in Table 4 clearly demonstrate that PatentGPT outperforms state-of-the-art LLMs across all metrics, including ChatGPT-4o, Gemini, and Claude2, in generating high-quality patentable text. Compared to ChatGPT-4o (86.899) and Gemini (86.541), the superior BERTScore suggests that PatentGPT excels in understanding the domain-specific terminology and conceptual depth required for patent writing. PatentGPT achieves a BLEU-4 score of 45.36, significantly outperforming the other baselines, which highlights PatentGPT's superior ability to generate highly precise and syntactically coherent patent content. At the same time, PatentGPT also has consistently higher ROUGE scores across all variants, demonstrating its ability to accurately and fluently generate text that closely mirrors the stylistic and structural elements of real patent documents.

### 5.2. Comparison results of different methods

To highlight the advantages of knowledge-based fine-tuning over traditional fine-tuning, we systematically compare the performance of various approaches. Table 5 and Fig. 4 provide a comprehensive view of how each training method impacts evaluation metrics. Specifically:

**KPT vs. PT**: The results show that while models trained with only KPT and PT perform similarly in terms of BERT Score and BLUE-4, the real advantage of KPT becomes evident when these models undergo further task-specific training. For example, the KPT model achieves a BERT Score of 82.902 and a BLUE-4 score of 22.015, which are relatively close to those of the PT model, which has a BERT Score of 79.651 and a BLUE-4 score of 11.663. This suggests that knowledge injection does not drastically alter base model performance without task-specific fine-tuning but provides a solid foundation for subsequent improvements.

**Impact of SFT on PT and KPT**: When SFT is applied, the models that underwent knowledge-injected pre-training (KPT + SFT) show a significantly larger performance boost compared to those that only received traditional pre-training (PT + SFT). The BERT Score for KPT + SFT is 88.305, and the BLUE-4 score reaches 44.661, both of which are significantly higher than the PT + SFT model's BERT Score of 86.621 and BLUE-4 score of 30.848. This dramatic improvement highlights the effectiveness of knowledge injection during pre-training, which helps the model better leverage the fine-tuning process to adapt to specific tasks.

**Combined Training with RLHF**: The final step combines KPT, SFT, and RLHF. The model trained with KPT + SFT + RLHF achieves the highest scores across all metrics, with a BERT Score of 90.003 and a BLUE-4 score of 45.360. These results are superior to those of the PT + SFT + RLHF model. This comprehensive approach, combining knowledge injection, task-specific fine-tuning, and human feedback, maximizes the model's ability to perform complex tasks.

Fig. 4 visually compares fine-tuning approaches across the five metrics: The left group in each cluster represents traditional finetuning methods (PT, PT + SFT, PT + SFT + RLHF). The right group corresponds to KFT-based methods (KPT, KPT + SFT, KPT + SFT + RLHF). By comparing the bar heights and labeled numeric scores in Fig. 4, one can observe that once the model is infused with domain knowledge (i.e., KPT), subsequent SFT and RLHF yield a more substantial improvement,

**Table 4**
Automated evaluation results.

| Model | Parameters Size | Bert Score | BLUE-4 | ROUGE-1 | ROUGE-2 | ROUGE-L |
|---|---|---|---|---|---|---|
| Qwen2 [30] | 1.5b | 78.396 | 9.705 | 18.640 | 5.812 | 7.589 |
| Llama2 [31] | 7b | 79.086 | 11.469 | 14.835 | 3.371 | 8.069 |
| Llama3.2 [32] | 3b | 78.754 | 8.556 | 14.940 | 4.031 | 7.615 |
| phi-2 [32] | 2.7b | 81.496 | 15.799 | 17.270 | 5.263 | 10.96 |
| ChatGPT-4o [33] | 8*220b | 86.899 | 17.232 | 38.152 | 14.216 | 31.946 |
| Gemini [34] | 8b | 86.541 | 16.721 | 38.133 | 15.677 | 32.767 |
| Claude2 [35] | 137b | 83.727 | 11.022 | 32.545 | 12.414 | 28.147 |
| MoZI [25] | 7b | 74.517 | 8.663 | 12.522 | 2.366 | 6.318 |
| PatentGPT | 1.5b | 90.003 | 45.360 | 41.961 | 25.977 | 36.022 |

**Table 5**
Comparison of methods at different stages.

| Method | Bert Score | BLUE-4 | ROUGE-1 | ROUGE-2 | ROUGE-L |
|---|---|---|---|---|---|
| PT | 79.651 | 11.663 | 19.452 | 5.609 | 8.900 |
| PT + SFT | 86.625 | 30.848 | 36.503 | 18.550 | 27.813 |
| PT + SFT + RLHF | 83.459 | 12.015 | 24.885 | 12.186 | 16.609 |
| KPT | 82.902 | 22.015 | 18.908 | 4.728 | 12.866 |
| KPT + SFT | 88.305 | 44.661 | 41.341 | 25.271 | 35.259 |
| KPT + SFT + RLHF | 90.003 | 45.360 | 41.961 | 25.977 | 36.022 |

aligning with the quantitative findings reported in Table 5.

### 5.3. Ablation study

To validate the contributions of each module in our proposed KFT framework, we conducted an ablation study systematically removing key training stages (KPT, SFT, RLHF) and assessing their impacts on the quality of generated patent texts. As shown in Fig. 5 and Table 6, when RLHF is omitted, a minor decline in metrics is observed reflecting RLHF's role in refining outputs to align with human expectations. Removing both SFT and RLHF results in a significant decline, with BLEU-4 dropping by 51.4 % and ROUGE-1 by 54.9 %, emphasizing SFT's critical role in equipping the model with task-specific knowledge. Moreover, removing KPT severely deteriorates the model's performance, with BLEU-4 decreasing by 78.6 % and ROUGE-2 by 77.6 %. These results demonstrate that KPT serves as the indispensable foundation providing the specialized domain knowledge necessary for patent generation.

Additionally, Fig. 5 visualizes the impact of each training component on model performance, clearly illustrating how the progressive integration of KPT, SFT, and RLHF enhances overall quality. Thus, our ablation study confirms that each step of our KFT framework significantly contributes to the model's capability to generate high-quality patent texts.

### 5.4. Patent bench

The Patent Bench results (Table 7) highlight the performance of different models across four distinct evaluation tasks: IP Quiz, IP Exam, IP Match, and MMLU. Overall, PatentGPT exhibits superior performance across most benchmarks, as shown in Fig. 6, particularly excelling in tasks that require specific IP-related knowledge. The results also show that PatentGPT's score in MMLU is relatively low compared to other large parameter models (e.g., ChatGPT-4o and Claude2). However, from the perspective of the model itself, compared to the un-fine-tuned model of the same parameter size Qwen2, PatentGPT sacrifices a small part of its general capabilities, but it greatly improves its professional capabilities in the IP field aligning with its design goals of generating patent concepts. In summary, PatentGPT achieves a significant boost in specialized IP tasks while maintaining competitive general capabilities. Its smaller parameter size highlights its adaptability for targeted applications, making it a highly effective artifact for IP-related tasks.

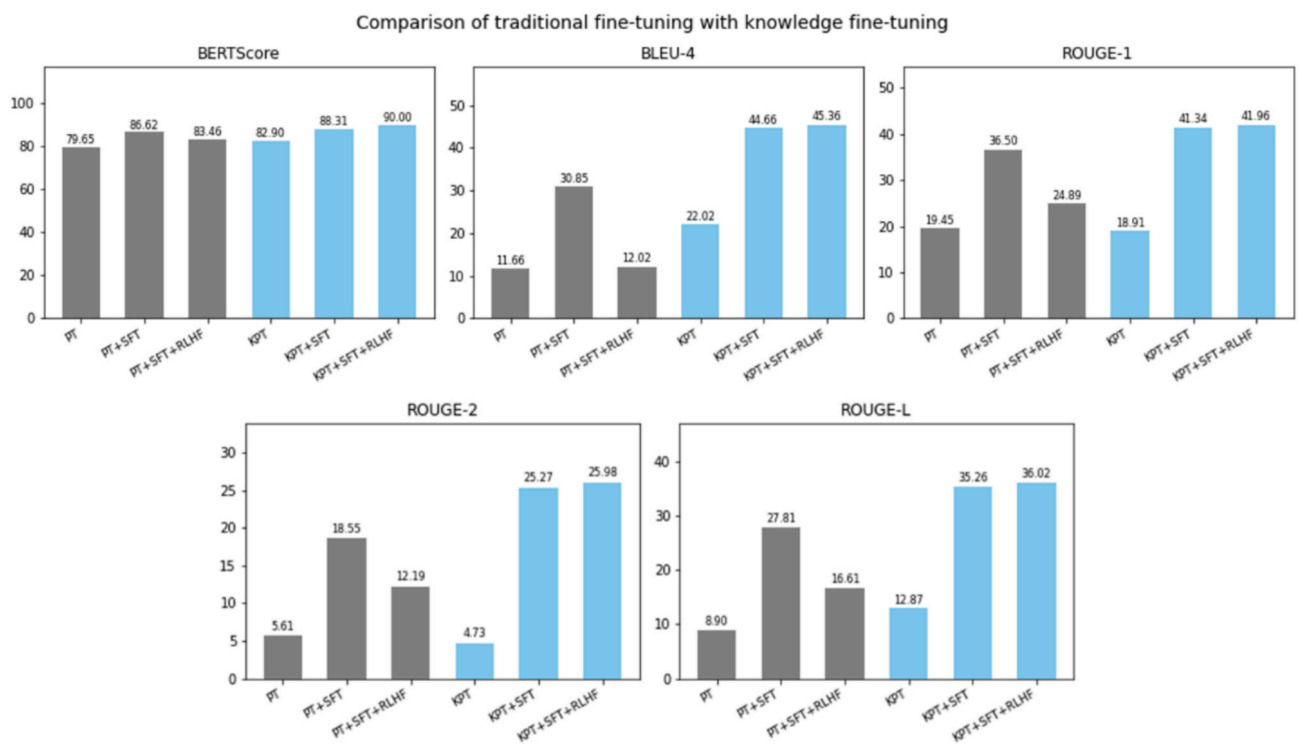

**Fig. 4.** Visual effect improvement of methods at different stages.

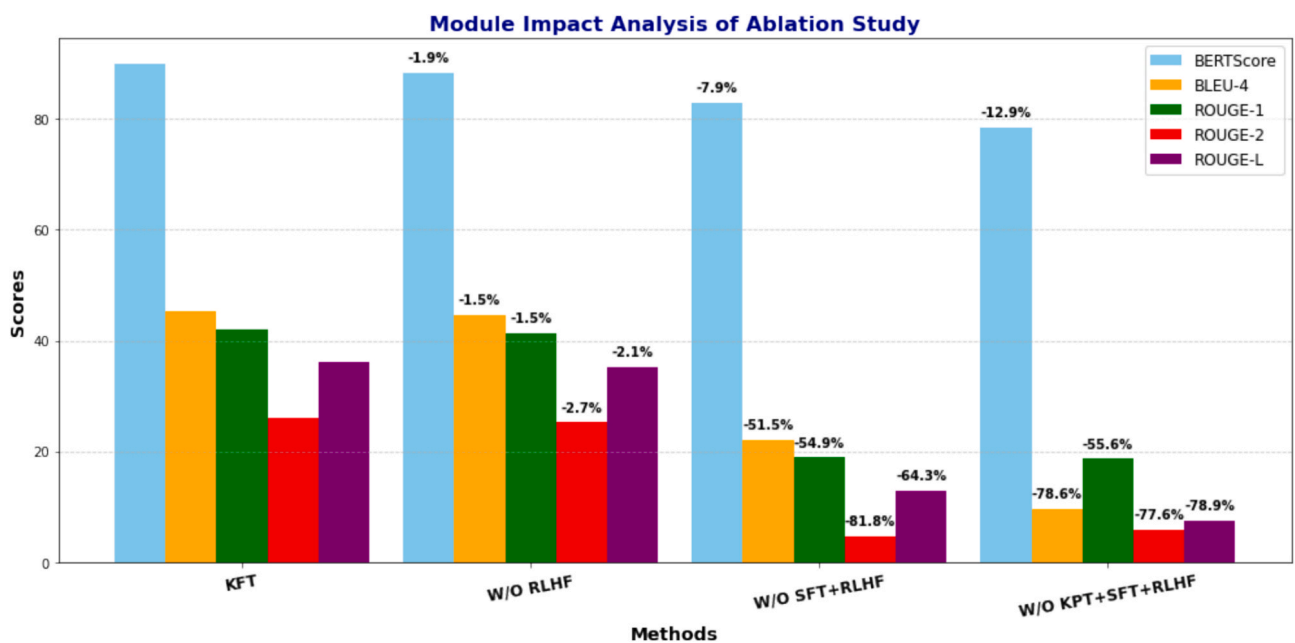

**Fig. 5.** Performance change trends at different training stages.

**Table 6**
Ablation study results.

| Method | Bert Score | BLUE-4 | ROUGE-1 | ROUGE-2 | ROUGE-L |
| --- | --- | --- | --- | --- | --- |
| KFT | 90.003 | 45.360 | 41.961 | 25.977 | 36.022 |
| W/O RLHF | 88.305 (−1.88 %) | 44.661 (−1.54 %) | 41.341 (−1.48 %) | 25.271 (−2.71 %) | 35.259 (−2.12 %) |
| W/O SFT + RLHF | 82.902 (−7.89 %) | 22.015 (−51.44 %) | 18.908 (−54.94 %) | 4.728 (−81.81 %) | 12.866 (−64.27 %) |
| W/O KPT + SFT + RLHF | 78.396 (−12.90 %) | 9.705 (−78.61 %) | 18.640 (−55.58 %) | 5.812 (−77.63 %) | 7.589 (−78.93 %) |

### 5.5. Reasonability and innovation

In this section, we evaluate the performance of different models on the benchmark tests of reasonability and innovation in concept generation. Reasonability is assessed by the conformity of generated texts to the USPTO standards, while innovation is quantified using the *Average Rareness Score*, which measures the novelty of concepts against the existing patent database. Reasonability refers to the model's ability to generate patentable concepts that conform to the USPTO standards. This is quantified by *Win* and *Loss*, indicating whether the model meets the writing standards and format. For the result of reasonability, PantetGPT has a 95 % win rate as shown in Fig. 7, which is significantly higher than all other baseline models.

The concept innovation was evaluated using the Average Rareness Score, which measures the novelty of generated concepts by calculating their knowledge distance from existing concept combinations in patents. A higher score indicates a greater degree of innovation. As shown in Fig. 8, models like MoZI and Qwen2 scored 0.402 and 0.431, respectively, reflecting their reliance on existing knowledge with limited ability to synthesize novel combinations. In contrast, PatentGPT achieved the highest average rareness score of 0.578, demonstrating its exceptional capability for generating innovative patent concepts. By KPT on concepts derived from existing knowledge of patents, PatentGPT develops a strong foundation of domain knowledge and concepts. The subsequent SFT enables it to focus on generating task-specific outputs, further enhanced by RLHF to align its results with human expectations of novelty and relevance. This multi-stage training process enables PatentGPT to balance creativity with practicality, generating concepts that are both innovative and applicable. Therefore, this experiment also demonstrated the effectiveness of the proposed KFT method from the perspective of innovative ontology in design.

### 5.6. Case study

We conduct a detailed case study comparing PatentGPT with the best-performing baseline model (i.e., ChatGPT-4o) in our experimental results. For the case study, we designed a realistic query based on a hybrid vehicle technology scenario, which required the models to generate a patent title, abstract, and claims. Table 8 and Table 9 respectively show the process of PatentGPT and ChatGPT-4o turning the same idea into patented text. The case study is structured to evaluate the outputs based on title, abstract, and claims aspects:

**Table 7**
Patent bench results.

| Type | PatentGPT | MoZI | Qwen2 | Llama2 | phi-2 | ChatGPT-4o | Gemini | Claude2 | Llama3.2 |
| --- | --- | --- | --- | --- | --- | --- | --- | --- | --- |
| IP_Quiz | 0.524 | 0.374 | 0.516 | 0.467 | 0.503 | 0.508 | 0.497 | 0.468 | 0.506 |
| IP_Exam | 0.33 | 0.170 | 0.33 | 0.24 | 0.24 | 0.36 | 0.30 | 0.317 | 0.30 |
| IP_Match | 0.262 | 0.258 | 0.254 | 0.260 | 0.24 | 0.384 | 0.351 | 0.347 | 0.348 |
| MMLU | 0.535 | 0.316 | 0.55 | 0.464 | 0.533 | 0.887 | 0.789 | 0.785 | 0.605 |

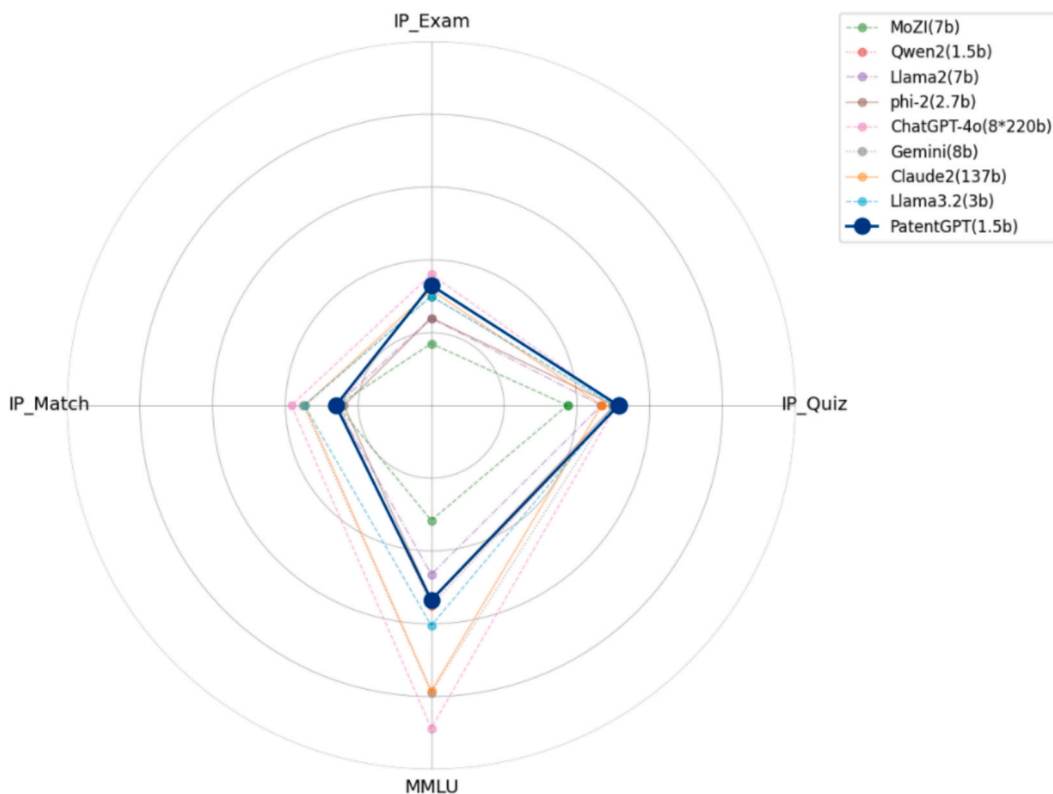

**Fig. 6.** Patent bench results.

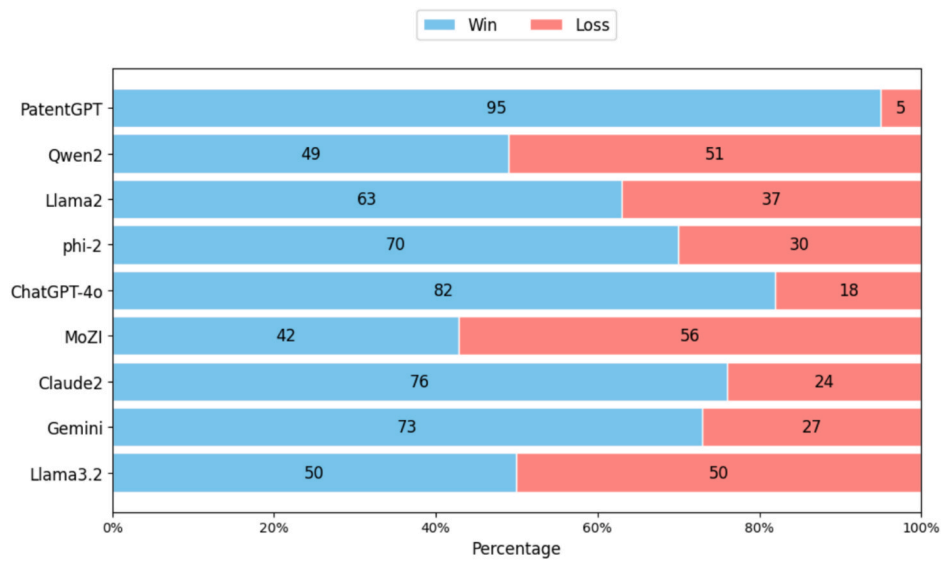

**Fig. 7.** Reasonability results.

**Title**: The title of PatentGPT exhibits higher specificity and contextual relevance by clearly identifying the system's structure ("engine-transmission connection system") and function ("control method"). In contrast, ChatGPT-4o's title is broader and lacks structural specificity, which may hinder its ability to define the patent scope adequately. From a linguistic perspective, the generated title of PatentGPT avoids redundancy and aligns well with industry-specific terminology.

**Abstract**: The abstract of PatentGPT provides a technically comprehensive and application-oriented description. By explicitly mentioning operational details, such as "dynamic management of power distribution," it positions the invention within a practical context, increasing its relevance for industry professionals and patent examiners. In contrast, the generated of ChatGPT-4o, while coherent, relies on broader descriptions that might fail to fully convey the invention's novelty and applicability. Hence, PatentGPT's ability to integrate practical utility with technical depth demonstrates a higher degree of adaptability to real-world use cases.

**Claims**: PatentGPT generated a robust set of claims, featuring:

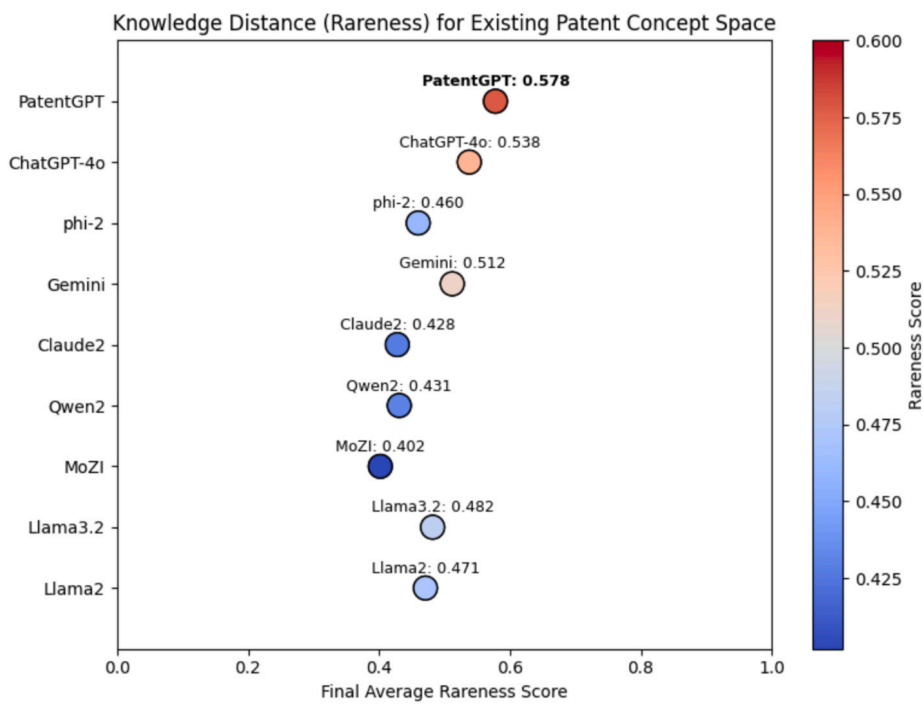

**Fig. 8.** Concept innovation distance visualization.

Independent claims defining the system and method. Dependent claims elaborating on specific features, such as regenerative braking, load balancing, and power transitions. A well-organized hierarchical structure ensuring clear coverage of all technical aspects. In comparison, ChatGPT-4o's claims were functional but less comprehensive. For example, they omitted detailed scenarios for managing driving mode transitions and regenerative braking, limiting their legal robustness and technical applicability. This difference underscores PatentGPT's ability to produce legally defensible claims that comprehensively define the invention's scope.

## 6. Discussion

### 6.1. Implications for Practice and Governance

The development of PatentGPT through KFT presents significant methodological contributions to engineering and technology management. After the benchmark testing, this approach demonstrates that leveraging LLMs with domain-specific knowledge injections can enhance their capacity to handle specialized, technical tasks such as initial patent drafting. The KFT methodology suggests a shift towards integrating domain knowledge directly into the model's training process, which contrasts with traditional LMs and LLMs that rely heavily on general-purpose data. However, the application of AI-generated patents also introduces legal, economic, and ethical considerations that must be addressed.

One of the most contentious legal issues surrounding AI in intellectual property law is whether AI-generated inventions qualify for patent protection and whether an AI system itself can be recognized as an inventor. Recent legal cases have led to divergent rulings of jurisdictions, reflecting the ongoing global debate on AI inventorship. The USPTO and subsequent court rulings in Thaler v Commissioner of Patents upheld that the U.S. patent system requires a natural person as an inventor [44,45]. This stance reflects a broader legal philosophy that prioritizes human creativity and accountability, ensuring that patent rights are granted to entities capable of moral and legal responsibility. The European Patent Office (EPO) similarly rejected Dr. Thaler's DABUS patent applications, stating that patent law requires human inventors. This position aligns with the EU's broader emphasis on human-centric innovation and ethical AI development. However, critics argue that this rigid interpretation may stifle innovation by failing to recognize the growing role of AI in the inventive process [46]. The UKIPO and High Court of Appeal upheld the rejection of DABUS as an inventor, ruling that the Patents Act 1977 requires a natural person as an inventor [47]. Notably, the UK courts have suggested that legislative reform would be necessary to accommodate AI-generated inventions, reflecting a cautious approach to redefining traditional patent law frameworks. This highlights a potential pathway for jurisdictions to adapt their laws while maintaining core principles of human accountability. In contrast, the Federal Court of Australia ruled that AI could be listed as an inventor, stating that the word "inventor" should be interpreted broadly and that AI-generated inventions could be assigned to a human applicant [48]. However, this ruling was later overturned on appeal. This reversal underscores the judicial reluctance to expand inventorship beyond natural persons without explicit legislative mandate, reflecting concerns about legal consistency and the potential erosion of human-centric patent systems. In 2021, the South African Patent Office approved a patent application for artificial intelligence as an inventor [49]. This decision, however, has been criticized for lacking substantive legal reasoning and is seen as an outlier rather than a precedent-setting move. These cases highlight the lack of global consensus on AI inventorship. Moving forward, legislative bodies will need to revisit patent laws to address the growing role of AI in innovation.

From an economic perspective, the integration of AI into the patenting process has the potential to democratize innovation. By reducing costs and streamlining the drafting process, PatentGPT can make the patent system more accessible to smaller entities and individual inventors, fostering a more inclusive innovation ecosystem. This could accelerate innovation cycles and drive economic growth, particularly in high-tech industries. However, it also raises concerns about job displacement, particularly in roles traditionally associated with patent drafting and review [50].

Ethical implications are also central to this discourse. As AI systems increasingly contribute to creative processes, questions arise regarding the originality and authenticity of their outputs. If AI outpaces human creativity in certain domains, how should society value these outputs in

**Table 8**
The case of PatentGPT.

*Query*
I have the following ideas:
Explore the technology of connecting the engine and transmission through a torque converter, and design a control method to achieve efficient collaboration between the electric motor and the internal combustion engine.
Could you please help me come up with a suitable patent title based on my ideas for application?

*Answer*
Torque Converter-Based Engine-Transmission Connection System and Control Method for Efficient Collaboration Between Electric Motor and Internal Combustion Engine. This title captures the key elements of your idea: the use of a torque converter to connect the engine and transmission, and the design of a control method to enable efficient collaboration between the electric motor and the internal combustion engine. Let me know if you'd like to refine it further!

*Query*
Here is the title of the patent I want to apply for: Torque Converter-Based Engine-Transmission Connection System and Control Method for Efficient Collaboration Between Electric Motor and Internal Combustion Engine. Could you please write a corresponding abstract?

*Answer*Abstract: A torque converter-based system and control method are disclosed for connecting an internal combustion engine and a transmission to enable efficient collaboration with an electric motor in hybrid or electric vehicles. The system includes a torque converter configured to optimize power transfer between the engine and the transmission, an electric motor connected to the transmission and capable of operating independently or in conjunction with the engine, and a control unit designed to coordinate their operations. The control method dynamically manages power distribution, load balancing, and energy efficiency by adjusting torque and speed based on vehicle demands and operating conditions. This invention enhances system performance, fuel efficiency, and overall driving experience while reducing emissions.

*Query*
Could you please write the corresponding claim text based on my title and abstract?

*Answer*
Claims:1. A torque converter-based system for connecting an internal combustion engine and a transmission in a vehicle, comprising:

o a torque converter configured to facilitate power transfer between the internal combustion engine and the transmission;
o an electric motor connected to the transmission, wherein the electric motor is operable independently or in conjunction with the internal combustion engine; and
o a control unit configured to coordinate operations of the internal combustion engine and the electric motor to optimize power distribution, torque output, and energy efficiency based on vehicle demands and operating conditions.

2. The system of claim 1, wherein the torque converter is further configured to balance torque between the internal combustion engine and the transmission during acceleration and deceleration events.
3. The system of claim 1, wherein the control unit dynamically adjusts the operation of the electric motor and the internal combustion engine to minimize fuel consumption and reduce emissions.
4. The system of claim 1, wherein the electric motor is connected to the transmission through a planetary gear system.5. A method for operating a torque converter-based system for a vehicle, comprising:

o transferring power from an internal combustion engine to a transmission through a torque converter;
o operating an electric motor independently or in conjunction with the internal combustion engine to provide additional torque or operate the vehicle under specific conditions; and
o dynamically managing power distribution and energy efficiency using a control unit that coordinates the operation of the internal combustion engine and the electric motor based on real-time vehicle demands.

6. The method of claim 5, further comprising:

o balancing torque output between the internal combustion engine and the electric motor during transitions between different driving modes.

7. The method of claim 5, wherein the control unit calculates optimal torque and speed for the internal combustion engine and the electric motor to enhance vehicle performance and fuel efficiency.
8. The method of claim 5, further comprising using regenerative braking to recover energy and store it in a battery connected to the electric motor.9. The system of claim 1, wherein the control unit is further configured to switch between pure electric, hybrid, and combustion engine modes based on predefined driving conditions.10. The method of claim 5, wherein the control unit monitors external conditions, including load, gradient, and speed, to determine the appropriate collaboration strategy between the internal combustion engine and the electric motor.

---

legal and professional contexts? Organizations such as the World Intellectual Property Organization (WIPO) have initiated discussions on these topics, reflecting a growing recognition of the need to address how AI-generated content aligns with global norms around intellectual property [51]. For example, the European Union is also exploring legislative reforms to address AI-generated IP, with proposals aiming to clarify the role of AI in innovation and ensure accountability mechanisms are in place [52].

In response to these developments, it is crucial for both organizations and nations to formulate correspondent technology policies to prevent inequality. In conclusion, as we stand at the intersection of technology and management, the path forward should involve a balanced approach to AI policy and regulation, fostering an environment where AI supports human efforts in creative processes without undermining the essential human elements that drive innovation. This balance is essential for maintaining public trust and maximizing the benefits of AI in invention generation. It should ensure that the patent management method continues to fulfill its role in promoting human ingenuity while adapting to new technological realities.

### 6.2. Limitation and future work

While PatentGPT demonstrates strong capabilities in generating high-quality patent texts, several limitations remain. These limitations can be categorized into the model's domain adaptability, novelty and legal compliance, and potential impact on the patent ecosystem:

(1) **Adaptability to Jurisdictional Variations**. Patent laws and drafting conventions vary widely across jurisdictions, such as claim structure, count restrictions, and procedural norms, which may result in PatentGPT producing text that fails to comply with international patent standards. PatentGPT's training data primarily consists of USPTO patents, which limits its applicability to generating jurisdiction-specific claims tailored to regional legal requirements. To address this limitation, future work will explore retrieval-augmented generation (RAG) and jurisdiction-specific prompt engineering, integrating legal guidelines and predefined templates from different patent offices (e.g., EPO, JPO, CNIPA). Additionally, we plan to incorporate cross-jurisdictional patent datasets to optimize PatentGPT for multi-regional patent drafting.

(2) **Applicability to Emerging Domains**. PatentGPT's current training data is based on patents from 2021 to 2022, meaning it may lack comprehensive knowledge of emerging fields, such as quantum computing, brain-computer interfaces, and synthetic biology. Without external retrieval mechanisms or sufficient instructions, PatentGPT may generate hallucinated or technically inaccurate claims in these areas. To improve its adaptability, future work will explore continuous knowledge updates by integrating emerging patent data.

(3) **Novelty, Legal Compliance, and Risk of Plagiarism**. Although

**Table 9**
The case of ChatGPT-4o.

*Query*
I have the following ideas:
Explore the technology of connecting the engine and transmission through a torque converter, and design a control method to achieve efficient collaboration between the electric motor and the internal combustion engine.
Could you please help me come up with a suitable patent title based on my ideas for application?

*Answer*
Based on your ideas, here is a possible patent title:" Control System and Method for Optimizing Collaboration Between Electric Motor and Internal Combustion Engine via Torque Converter in Hybrid Powertrains". Let me know if you'd like further refinements!

*Query*
Here is the title of the patent I want to apply for: Control System and Method for Optimizing Collaboration Between Electric Motor and Internal Combustion Engine via Torque Converter in Hybrid Powertrains. Could you please write a corresponding abstract?

*Answer*Abstract: The present invention relates to a control system and method for optimizing the collaboration between an electric motor and an internal combustion engine in a hybrid powertrain, connected via a torque converter. The system is designed to enhance energy efficiency, performance, and responsiveness by precisely managing power distribution and torque delivery during various operating conditions. The control method employs real-time monitoring of key parameters, including engine speed, motor speed, torque demands, and vehicle load, to dynamically adjust the interaction between the electric motor and the internal combustion engine. The torque converter acts as an adaptive mediator, ensuring smooth transitions and efficient energy utilization during power mode shifts. This innovation provides a robust solution for improving hybrid vehicle performance, reducing fuel consumption, and minimizing emissions while maintaining driving comfort and reliability. Let me know if adjustments are needed!

*Query*
Could you please write the corresponding claim text based on my title and abstract?

*Answer*
Claims:
1. Control system for a hybrid powertrain, comprising:

o an internal combustion engine;
o an electric motor;
o a torque converter operatively connecting the internal combustion engine and the electric motor; and
o a control unit configured to:
- monitor real-time parameters including engine speed, motor speed, torque demand, and vehicle load;
- dynamically adjust power distribution between the internal combustion engine and the electric motor; and
- control the torque converter to optimize torque delivery and ensure smooth transitions between power sources.

2. The control system of claim 1, wherein the control unit employs an adaptive algorithm to adjust the interaction between the internal combustion engine and the electric motor based on driving conditions, such as acceleration, deceleration, and cruising.3. The control system of claim 1, wherein the torque converter is configured to:

o mediate power mode shifts between the internal combustion engine and the electric motor;
o reduce transmission losses during torque delivery; and
o provide enhanced energy efficiency.

4. A method for optimizing collaboration in a hybrid powertrain, the method comprising:

o monitoring real-time parameters of a hybrid powertrain, including engine speed, motor speed, torque demand, and vehicle load;
o dynamically controlling power distribution between an internal combustion engine and an electric motor through a torque converter;
o adjusting torque delivery to achieve efficient energy utilization and smooth transitions during power mode changes.

5. The method of claim 4, wherein the dynamic control includes prioritizing electric motor operation during low-speed or stop-and-go driving conditions to reduce fuel consumption.6. The method of claim 4, further comprising:

o implementing predictive control algorithms to anticipate power demands based on driver input and environmental conditions; and
o pre-adjusting torque converter parameters to optimize energy efficiency.

7. The control system of claim 1, wherein the system is integrated into a hybrid vehicle to reduce emissions and improve fuel efficiency while maintaining driving comfort and reliability.

---

PatentGPT has been quantitatively evaluated for novelty using a rareness score, this does not replace the legal assessment of originality and patentability. The current evaluation framework lacks qualitative validation from patent attorneys and does not systematically assess compliance with patent laws across jurisdictions. Moreover, there remains a risk of unintentional plagiarism, as the model may generate claims that closely resemble existing patents without proper attribution. Future work will integrate legal expert evaluations to qualitatively assess patentability, sufficiency of disclosure, and compliance with regional laws.

(4) **Potential Misuse and Ethical Implications**. PatentGPT's ability to streamline patent drafting lowers the barrier for researchers and engineers applying for patents. However, this also introduces potential risks. Large corporations may leverage AI-generated patents to file excessive or low-quality patents, making it difficult for smaller companies and startups to have IP landscapes. Without proper oversight, AI-generated patents could flood patent offices, increasing the workload for patent examiners and reducing the overall quality of granted patents. To address these concerns, future work will explore AI auditing frameworks to detect and prevent unethical patent filing practices. We also plan to engage with patent offices and policymakers to study the broader impact of generative AI on intellectual property law and patent quality.

## 7. Summary

The PatentGPT can generate concepts that are patentable. And patent texts represent novel and useful concepts. Thus, the work is about the generation of concepts and patents together. The development of PatentGPT, through the innovative KFT framework, addresses a critical gap in the capabilities of current LLMs by equipping them with specialized knowledge essential for generating concepts and inventions. Our approach, which integrates KPT with domain-specific SFT and RLHF, significantly enhances the model's ability to comprehend and articulate complex technical concepts. The empirical results demonstrate that PatentGPT not only surpasses existing models on IP-related benchmarks but also sets a new benchmark for AI-driven IP creation.

### CRediT authorship contribution statement

**Runtao Ren:** Writing – review & editing, Writing – original draft, Visualization, Validation, Software, Methodology, Formal analysis, Data curation, Conceptualization. **Jian Ma:** Supervision, Funding acquisition, Conceptualization. **Jianxi Luo:** Writing – review & editing, Validation, Supervision.

**Declaration of competing interest**

The authors declare that they have no known competing financial interests or personal relationships that could have appeared to influence the work reported in this paper.

**Appendix A. Prompt for extracting knowledge graphs**

**Prompt of extracting knowledge graphs**

Goal:
Given a text document that is potentially relevant to this activity and a list of entity types, identify all entities of those types from the text and all relationships among the identified entities.
Entity Types:
1.Task: Applications, problems to solve, systems to construct
2.Method: Methods, models, systems to use, or tools, components of a system, frameworks
3.Material: Data, datasets, resources, corpus, knowledge base
4.Generic: General terms or pronouns that may refer to an entity but are not themselves informative, often used as connection words

Relation Types:
1.Used-for: B is used for A, B models A, A is trained on B, B exploits A, A is based on B
2.Feature-of: B belongs to A, B is a feature of A, B is under A domain
3.Hyponym-of: B is a hyponym of A, B is a type of A
4.Part-of: B is a part of A
5.Compare: Symmetric relation. Opposite of conjunction, compare two models/methods, or listing two opposing entities
6.Conjunction: Symmetric relation. Function as similar role or use/incorporate with

Step:
1. Identify all entities in the text. For each identified entity, extract the following information:
Entity_name: Name of the entity, capitalized
Entity_type: One of the following types: [Task, Method, Material, Generic]
Entity_description: A comprehensive description of the entity's attributes and activities

2. Identify all relationships among the entities from Step 1. For each clearly related pair of entities, extract:
Source_entity: name of the source entity (as identified in step 1)
Target_entity: name of the target entity (as identified in step 1)
Relationship_description: explanation why the source entity and the target entity are related
Telationship_strength: a numeric score indicating the strength of the relationship (e.g., 1-5)

3. Return the output in English as a single list of all identified entities and relationships, using {record_delimiter} as the delimiter for each record in the list.

4. When you have finished listing all entities and relationships, output on its own line.

**Appendix B. Prompt for generating knowledge corpus**

**Prompt of transforming knowledge**

You are now an expert in integrating knowledge. Please transform these interrelated knowledge into understandable concepts according to their relationships.

[Knowledge Graphs]

**Appendix C. Prompts for generating the text of patents**

**Prompt of generating titles**

I have the following ideas :

[Ideas]

Could you please help me come up with a suitable patent title based on my ideas for application?

**Prompt of generating abstracts**

Here is the title of the patent I want to apply for:

[Title]

Could you please write a corresponding abstract based on the title?

**Prompt of generating claims**

Here is the title and abstract of the patent I want to apply for:

[Title]

[Abstract]

Could you please write the corresponding claims?

**Appendix D. Learning curves for different training stages**

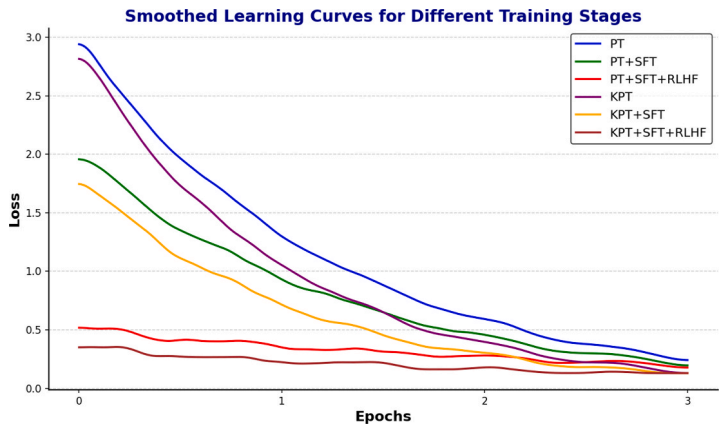

**Data availability**

Data will be made available on request.